\documentclass{article}

\usepackage{microtype}
\usepackage{graphicx}
\usepackage{subfigure}
\usepackage{booktabs} % for professional tables

\usepackage{hyperref}
\usepackage{natbib}

\usepackage[accepted]{icml2021}

\usepackage{times}
\usepackage{latexsym}
\usepackage{graphicx}
\usepackage{framed}

\usepackage[T1]{fontenc}
\usepackage[utf8]{inputenc}

\usepackage{microtype}

\usepackage{inconsolata}

\usepackage{soul}
\usepackage{enumitem}

\icmltitlerunning{Large Language Model (LLM) as a System of Multiple Expert Agents}

\begin{document}

\twocolumn[
\icmltitle{Large Language Model (LLM) as a System of Multiple Expert Agents: An Approach to solve the Abstraction and Reasoning Corpus (ARC) Challenge}



\icmlsetsymbol{equal}{*}

\begin{icmlauthorlist}
\icmlauthor{John Tan Chong Min}{nus}
\icmlauthor{Mehul Motani}{nus}
\end{icmlauthorlist}

\icmlaffiliation{nus}{Department of Electrical and Computer Engineering, National University of Singapore}

\icmlcorrespondingauthor{John Tan Chong Min}{johntancm@u.nus.edu}
\icmlcorrespondingauthor{Mehul Motani}{motani@u.nus.edu}

\icmlkeywords{Abstraction and Reasoning Corpus, Large Language Models, Abstraction Spaces}

\vskip 0.3in
]


\printAffiliationsAndNotice{} % otherwise use the standard text.

\begin{abstract}
We attempt to solve the Abstraction and Reasoning Corpus (ARC) Challenge using Large Language Models (LLMs) as a system of multiple expert agents. Using the flexibility of LLMs to be prompted to do various novel tasks using zero-shot, few-shot, context-grounded prompting, we explore the feasibility of using LLMs to solve the ARC Challenge. We firstly convert the input image into multiple suitable text-based abstraction spaces. We then utilise the associative power of LLMs to derive the input-output relationship and map this to actions in the form of a working program, similar to Voyager / Ghost in the MineCraft. In addition, we use iterative environmental feedback in order to guide LLMs to solve the task. Our proposed approach achieves 50 solves out of 111 training set problems (45\%) with just three abstraction spaces - grid, object and pixel - and we believe that with more abstraction spaces and learnable actions, we will be able to solve more. 
\end{abstract}

\section{Introduction}

\begin{figure}[t]
\centering
\includegraphics[width=0.5\textwidth]{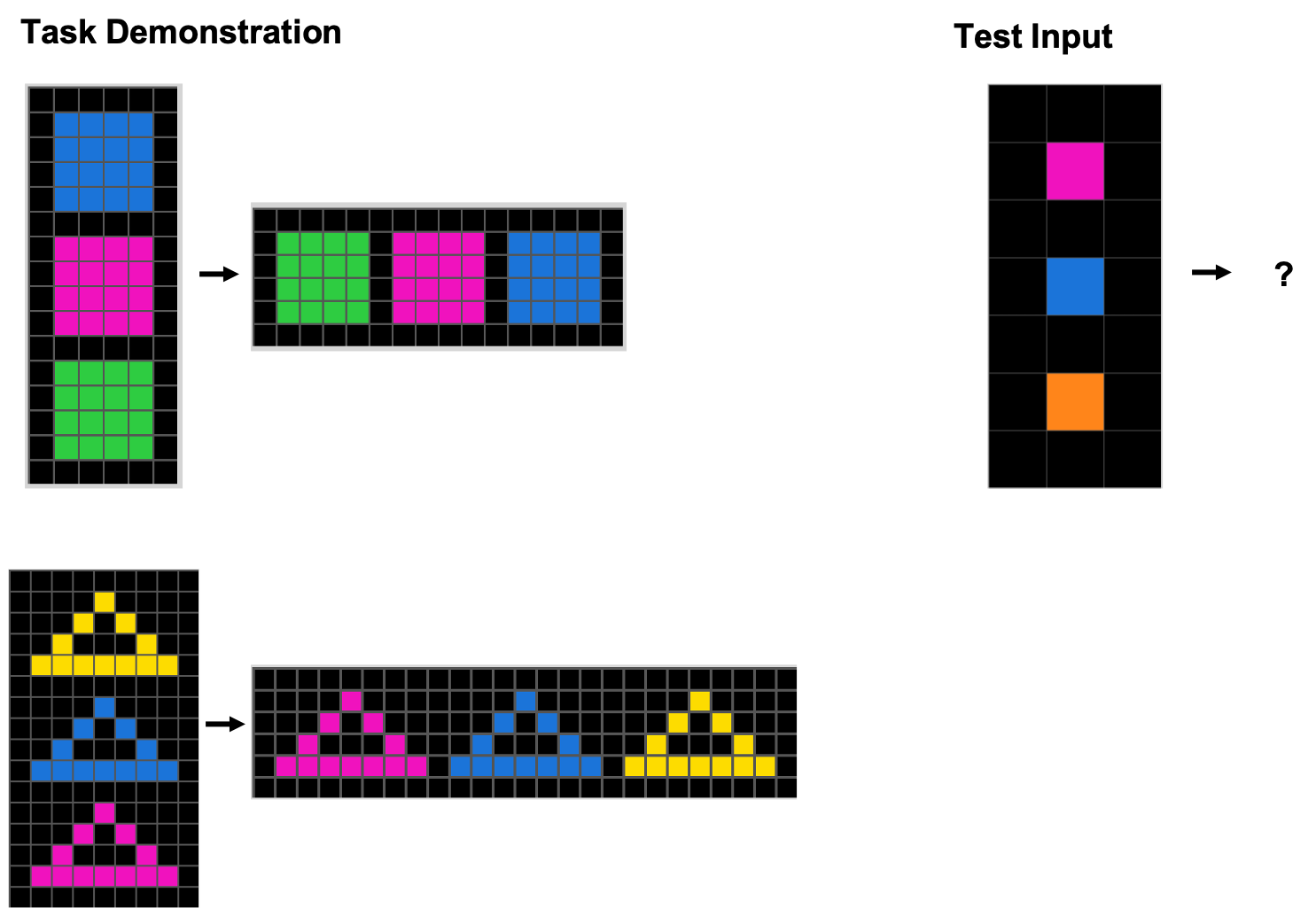}
\caption{A sample ARC task. The challenge is to infer the abstract rule(s) governing the demonstration transformations and apply it to the test input. Example from: \url{https://aiguide.substack.com/p/why-the-abstraction-and-reasoning}}
\label{ARC_example}
\vspace{-5mm}
\end{figure}

The Abstraction and Reasoning Corpus (ARC) Challenge is a key milestone in the march towards artificial general intelligence (AGI) as it requires forming concepts and abstractions \cite{chollet2019measure}. Fig. {\ref{ARC_example}} illustrates a sample ARC task. One of the key difficulties of the ARC challenge is that it requires doing something counter to mainstream deep learning – learning from very few samples. Deep learning typically uses tens of thousands of samples to do well. Humans, in comparison, can learn how to identify different animals by just one or two different observations. For instance, a child can identify a giraffe in real life for the first time, even though the only other time they may have been exposed to a giraffe was through a cartoon flash card. Such capabilities are not well endowed in modern AI systems, and that means that such AI systems will need to be trained extensively before deploying in the real world. After deploying them in the real world, they will also be limited in their ability to adapt and learn as the environment changes. 

In contrast, traditional symbol-based systems (e.g., GOFAI \cite{boden20144}) can “learn” quite fast, as any new situation can be interpreted without any learning phase, provided that there are existing symbols which can represent it. However, the history of GOFAI has shown that it is difficult to engineer these symbols, and at many times, even humans face difficulty to come up with symbols as they may not be able to express it in words.

As can be seen, there are shortcomings with the above two approaches, and a new kind of approach will be needed in order to learn fast and generalise to new situations, in order to even have a chance at solving the ARC Challenge. In this paper, we address this challenge by proposing to use Large Language Models (LLMs) as a system grounded in functional action spaces to tackle the ARC challenge. This can be said to be an intermediate ground between both deep learning and GOFAI approaches - The functional action spaces are more flexible than symbols in GOFAI; LLMs which are a form of deep learning that are adaptable to new situations via prompting. Specifically the contributions of the paper are as follows:
\begin{itemize}[nosep,leftmargin=*]
    \item We showcase a novel method of using LLMs as a system of multiple expert agents (without any pre-training) to solve the ARC Challenge
    \item We highlight the importance of a combination of multiple abstraction spaces from which to associate the input space to the output space
    \item We demonstrate the feasibility of grounding in functional space for program synthesis by LLMs.
\end{itemize}

\section{Related Work}
\begin{figure}[t]
\includegraphics[width=0.5\textwidth]{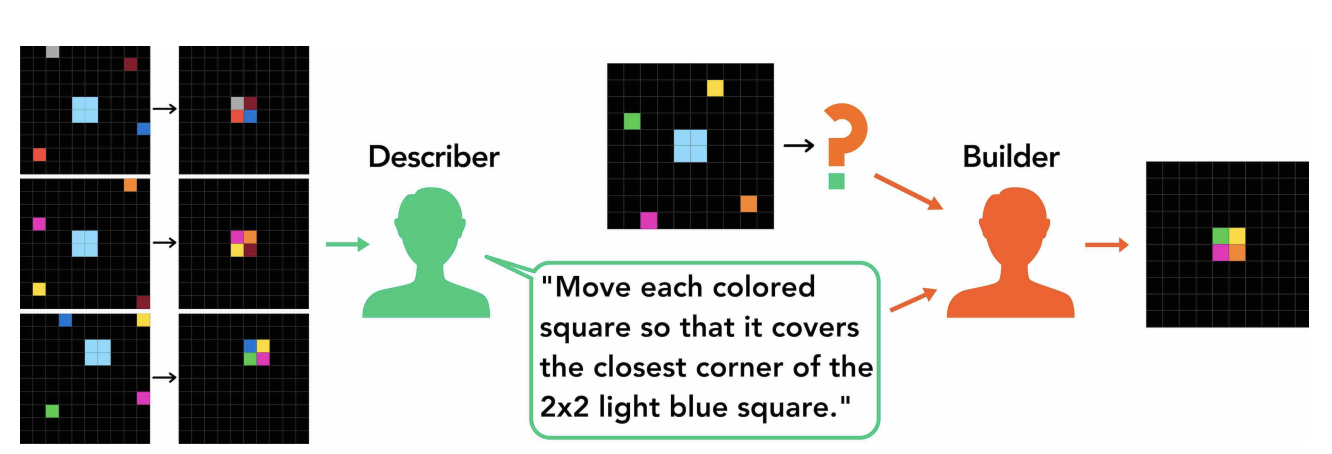}
\caption{88\% of ARC tasks can be solved by the Builder from just the description alone given by the Describer, without input-output examples. Can GPT-4 function as both the describer and the builder? Image reproduced from Fig. 4 of \citet{acquaviva2021communicating}.}
\label{fig:next_action_prediction}
% 	\caption{add caption here}
\label{fig: LARC}
\end{figure}

\textbf{ARC Challenge.} The ARC challenge \cite{chollet2019measure} comprises 400 public training tasks, 400 public evaluation tasks and 200 private test tasks. Each of these tasks has multiple "Task Demonstration" Input/Output grids, of which the task-taker must infer a common relation out of them. This common relation is then applied to the "Test Input", from which we get the "Test Output". The "Test Output" must match perfectly for it to be considered solved. The grids comprise grid sizes of 1x1 to 30x30, of which pixels can take on 10 different values.

\noindent \textbf{Domain Specific Language (DSL) Approaches.} The majority of ARC Challenge solutions are mainly DSL ones \cite{alford2021neurosymbolic, ferre2021first, xu2023graphs}. This is also the case for the first-place solution of the ARC Kaggle competition (\url{https://www.kaggle.com/code/icecuber/arc-1st-place-solution}).

\noindent \textbf{LLM-Based approaches.} One way to approach the ARC challenge will be to use text to describe the visual characteristics of objects \cite{camposampiero2023abstract}. Indeed, 88\% of ARC tasks can be solved via language description alone without input-output examples as shown in Fig. \ref{fig: LARC} \cite{acquaviva2021communicating}. For certain problems, denoting pixels in terms of objects can significantly boost the solve rate from 13 to 23 out of 50 object-related ARC tasks \cite{xu2023llms}. Some work has also been done to do end-to-end input to program description generation with just LLMs alone to some success \cite{min2023approach}. Other approaches have used Decision Transformers \cite{chen2021decision} to find a sequence of primitive actions from the input to output \cite{park2023unraveling}, however, as noted by the authors, huge amounts of data (10000 training data for 2000 testing data) are needed to train this method, it is unlikely it can generalise to unseen inputs. Recently, LLMs have been used to take the ASCII text view of the grid as input for next token prediction and have solved 85 out of 800 ARC tasks \cite{mirchandani2023large}.

\noindent \textbf{Code as Skills and Environmental Feedback.} Voyager is an embodied lifelong learning agent powered by LLMs \cite{wang2023voyager}. It features a skill library of functions to build up complex behaviour, and an iterative prompting mechanism with the environment to learn from environmental feedback. Ghost in the Minecraft \cite{zhu2023ghost} does something similar as well, though they constrain the action space to a list of functions. Similarly, we use code generation with primitive functions to approximate using a skill library, and use iterative prompting using ARC task output as feedback to learn from the environment.

\noindent \textbf{Our Method.} In line with the existing LLM approaches, we agree that we should use language as an alternate abstraction space in addition to the original pixel grid. Unlike existing approaches, we believe we should use more than one abstraction space. Hence, the LLM will be both the Builder and the Describer in Fig. \ref{fig: LARC}, but the Builder can also reference input-output pairs. We also believe we should integrate LLMs with a kind of DSL approach, but can afford to have an even more expressive DSL because an LLM is able to do matching of functions via semantics much more effectively than traditional DSL approaches.

\section{Broad Overview of Method}

\begin{figure*}[t]
\includegraphics[width=\textwidth]{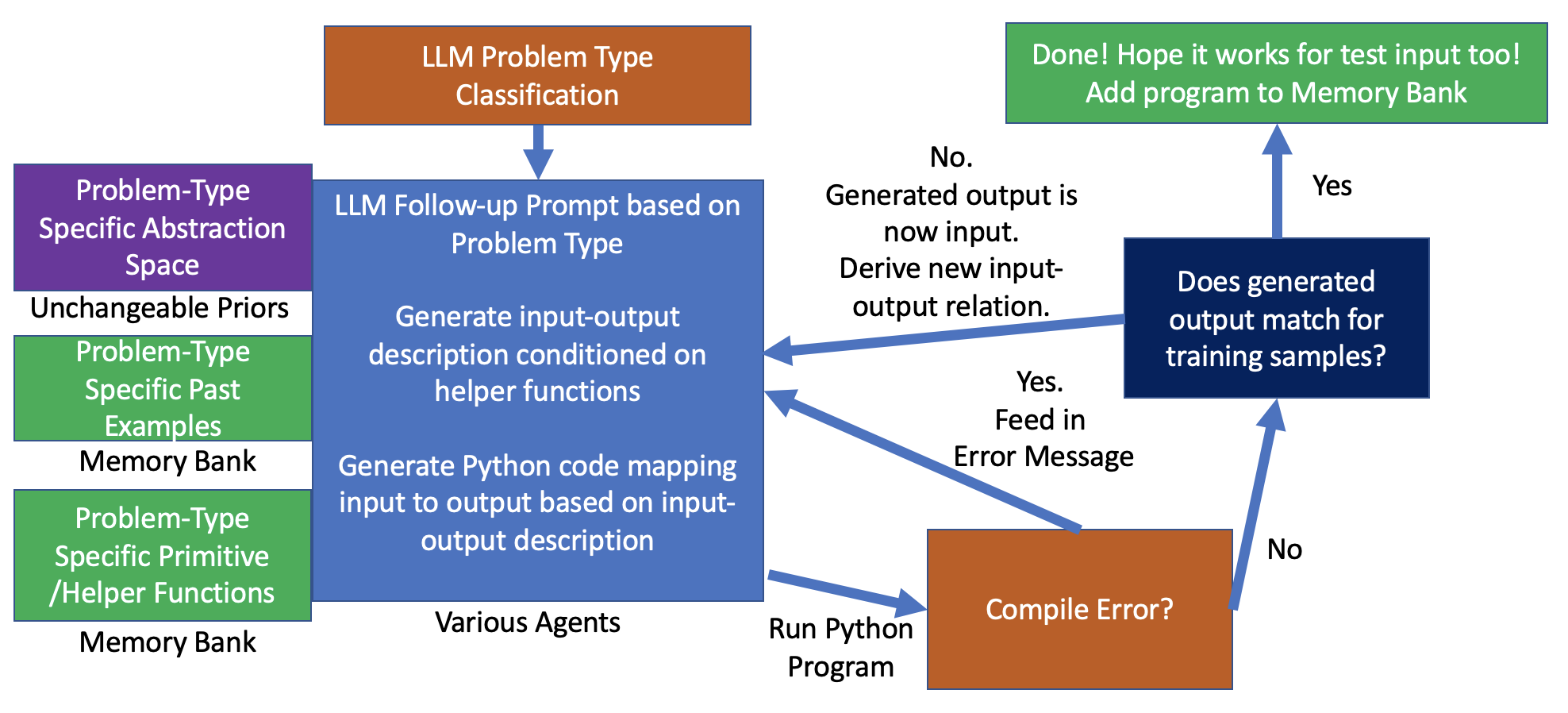}
\caption{Process Flowchart of LLMs as a System to solve the ARC Challenge.}
\label{fig: LLMs as System}
% 	\caption{add caption here}
\end{figure*}

In this section, we provide an overview of our proposed approach and discuss several key ideas behind it. We have not implemented out all parts of the proposed approach, but it is already doing well. Generative Pre-trained Transformer 4 (GPT-4) is a multimodal LLM created by OpenAI and released in March 2023 \cite{openai2023gpt4}. For now, we exclusively use GPT-4 for our model, as we empirically observe that GPT-3.5 and other open source models are not able to perform well enough for this method to work. The overall method is shown in Fig. \ref{fig: LLMs as System}.

\noindent \textbf{Problem Type Classification (Not Implemented).} ARC tasks test various concepts. If we can use past examples to ground the LLM, and let the LLM decide what problem category an ARC task belongs to, we can proceed with a specialised workflow to target solving that particular style of task. Presently, we simply run through all the various agent types and select the agent types which work. Implementing this classifier will not affect performance but will significantly help reduce the costs.

\noindent\textbf{Useful Abstraction Spaces.} While GPT-4 has proven to be a general purpose solver, being (currently) a text-based model, GPT-4 lacks some of the innate human priors necessary to solve the ARC challenge. For example, GPT-4 is not able to identify objects accurately from text alone. Objects are defined as continuous sections of the grid with the same non-zero value. Hence, providing such an object view as an abstraction space using text greatly helps with the GPT-4's ability to form associations with the input-output pair and is better able to find a solution \cite{xu2023llms}. Moreover, we can provide more than one abstraction space to GPT-4, which can increase the chance that one or more abstraction spaces contain a simple mapping from input to output, thereby reducing the complexity of the problem. Do note that these abstraction spaces are unchangeable, and are fixed since the beginning of learning. Hence, the agents will have to do processing based on these fixed priors.

\noindent\textbf{Encoding Human Biases via Helper/Primitive Functions.} An initial implementation of using GPT-4 to solve ARC was done with just prompting the human biases and action spaces via text. This did not do so well due to lack of grounding using words alone. A key innovation in this work is to use primitive functions as action spaces, as a way to encode human priors. If we could use functions for grounding, and express the semantic meaning of the function in words, GPT-4 could use the function to provide the code needed for the solution. Hence, the problem now becomes finding out what are the primitive functions we need to encode in order for the LLM to solve any generic ARC problem. 

\noindent\textbf{Using Memory for Additional Context (Not Implemented).} New problems might mix and match aspects of previous solutions, so having a memory bank to provide examples of similar solved problems in the past can help to ground the LLM to better generate the answer. This is currently not implemented due to constraints of context length. Once the context length for GPT-4 increases or fine-tuning becomes available, we intend to let each agent have memory of relevant previously solved problems and their solutions, so that it can ground the agent's output. This is akin to Retrieval Augmented Generation \cite{lewis2020retrieval}. 

\noindent\textbf{Utilising Feedback from Environment.} Another key idea is that a learning system would need to utilise feedback from the environment, and so a recursive loop feeding in feedback from the environment (whether there is compile error, whether the code matches the intended output) can help a lot in getting the right answer. This is akin to what is done in Voyager and Ghost in the MineCraft \cite{wang2023voyager, zhu2023ghost}.

\noindent\textbf{LLMs as a System.} Humans do not operate with only one system. We have various systems to call for various tasks. Similarly, we can have multiple expert agents for each task (such as \textit{Object View}, \textit{Pixel View}, \textit{Grid View}) and call on them to give their interpretation of the task, and select the most promising agent. This greatly helps narrow the search space for the solution. Then, we utilise the specialised functions this agent has and solve the problem. Interfacing this agent with environment feedback, the problem-type specific abstraction space, past examples and action spaces can greatly help filter and ground GPT-4 to generate a plausible solution. We believe that, with better grounding via expert agents, better abstraction space representations and better primitive function grounding, we will eventually be able to solve most of the ARC tasks using the proposed approach.

\section{Detailed Overview of Method}
We now go into some details of our method. Refer to Appendix \ref{appendix: prompt} and \ref{appendix: helper functions} for the full GPT-4 prompt.

\subsection{Different Abstraction Spaces}
We utilise various ways of encoding the abstraction spaces so that GPT-4 can better associate between the Input-Output pairs. It has been shown in Image-Joint Embedding Predictive Architecture (I-JEPA) \cite{assran2023self} and Stable Diffusion \cite{rombach2022high} that prediction in the latent/abstraction space leads to better downstream tasks than predicting in the input space. However, instead of just one abstraction space, we believe that there are many possible abstraction spaces which are fixed, and it is up to the solver to choose which is the best for the task at hand. We believe by incorporating more useful views and refining current ones, we can solve more ARC tasks.

For our method, we use only three views - \textit{Grid View}, \textit{Object View}, \textit{Pixel View }- and that has already achieved quite good results. In brief, \textit{Grid View} provides the entire grid representation, except we change the pixel numbers to characters so that we do not bias GPT-4 to treat it as an arithmetic problem to perform arithmetic on the pixel values. This also has the added benefit of ensuring that GPT-4 has not seen the ARC tasks before as it is now of a different form. The \textit{Object View} groups pixels that are contiguous together, so that they can be manipulated as a group. \textit{Pixel View} gives the coordinates for each pixel, which can help with more fine-grained movement tasks or relational tasks between pixels. Refer to Appendix \ref{appendix: abstraction} for more details. 

\subsection{JSON-based output format} LLMs are well known for being verbose and also relatively free-form in the output, making it hard for any automated program to use it. Here, we explicitly ask GPT-4 to output in a JSON format via prompting. This JSON format also facilities Chain-of-Thought (CoT) prompting \cite{wei2022chain}, as it is done in a specific sequence to encourage broad to specific thinking.

\subsection{CoT Prompting}
CoT enables the output to be structured and the LLM will be able to condition the generation of the later output based on the earlier ones. This enables a more broad to specific style of prompting, helping the LLM to think and reflect on various areas, narrowing the search space, and ultimately may help to solve the problem.

Here, we do CoT prompting directly using JSON format (See Appendix \ref{appendix: output examples} for some examples of GPT output in this JSON format). We ask GPT-4 to output: 
\begin{enumerate}[nosep,leftmargin=*]
\item "reflection": "reflect on the answer",     
\item "pixel\_changes": "describe the changes between the input and output pixels, focusing on movement or pattern changes",
\item "object\_changes": "describe the changes between the input and output objects, focusing on movement, object number, size, shape, position, value, cell count",
\item "helper\_functions": "list any relevant helper\_functions for this task",
\item "overall\_pattern": "describe the simplest input-output relationship for all input-output pairs",
\item "program\_instructions": "Plan how to write the python function and what helper functions and conditions to use",
\item "python\_program": "Python function named 'transform\_grid' that takes in a 2D grid and generates a 2D grid. Output as a string in a single line with $\backslash$n and $\backslash$t."
\end{enumerate}

\subsection{Helper/Primitive Functions}
For the functions, we basically zero-shot prompt by stating the function name plus the input parameters and the description of the function. We find that this format of zero-shot prompting works very well for most functions, especially if the name of the function is already indicative of what it does. This is very similar to the approach taken in Visual ChatGPT \cite{wu2023visual}, as well as OpenAI Functions (\url{https://openai.com/blog/function-calling-and-other-api-updates}). As this method of prompting is not sufficient to imbue biases that are not inherent in text (i.e. rotation, flipping), we also provide one-shot examples of how to use the function. 

\newpage
\subsection{Conditional Functions:} Rather than letting GPT-4 free-form generate its own code, we ask it to generate a conditional flow on the primitive functions. This greatly helps to reduce compilation errors. Such a conditional flow is needed, as some ARC tasks require using logic that only applies if a particular condition is met (e.g., turn the shape red if it has exactly 6 cells). Without this conditional flow, the program would need many more steps before it can solve the problem. An example of such a conditional flow is:\\
\textit{If \{condition\}: \{Primitive Function\}}

\section{Methodology}

\textbf{Select Problems by Context Length.} We firstly filter the ARC training set problems to only those whose Grid View and Object View (mono-color, no diagonals) can fit into a context length of 3000 tokens. This is important because later when we incorporate environmental feedback, we will need additional token length, and by empirical observation, 3000 tokens is necessary to guarantee some buffer token amount so that the entire prompt can fit within 8000 tokens later. This is the current maximum context length for the GPT-4 web browser, as well as for the basic GPT-4 API. In the future, we envision that our approach can work for more ARC tasks when the context length for GPT-4 increases.

\begin{figure}[t]
\centering
\includegraphics[width=0.5\textwidth]{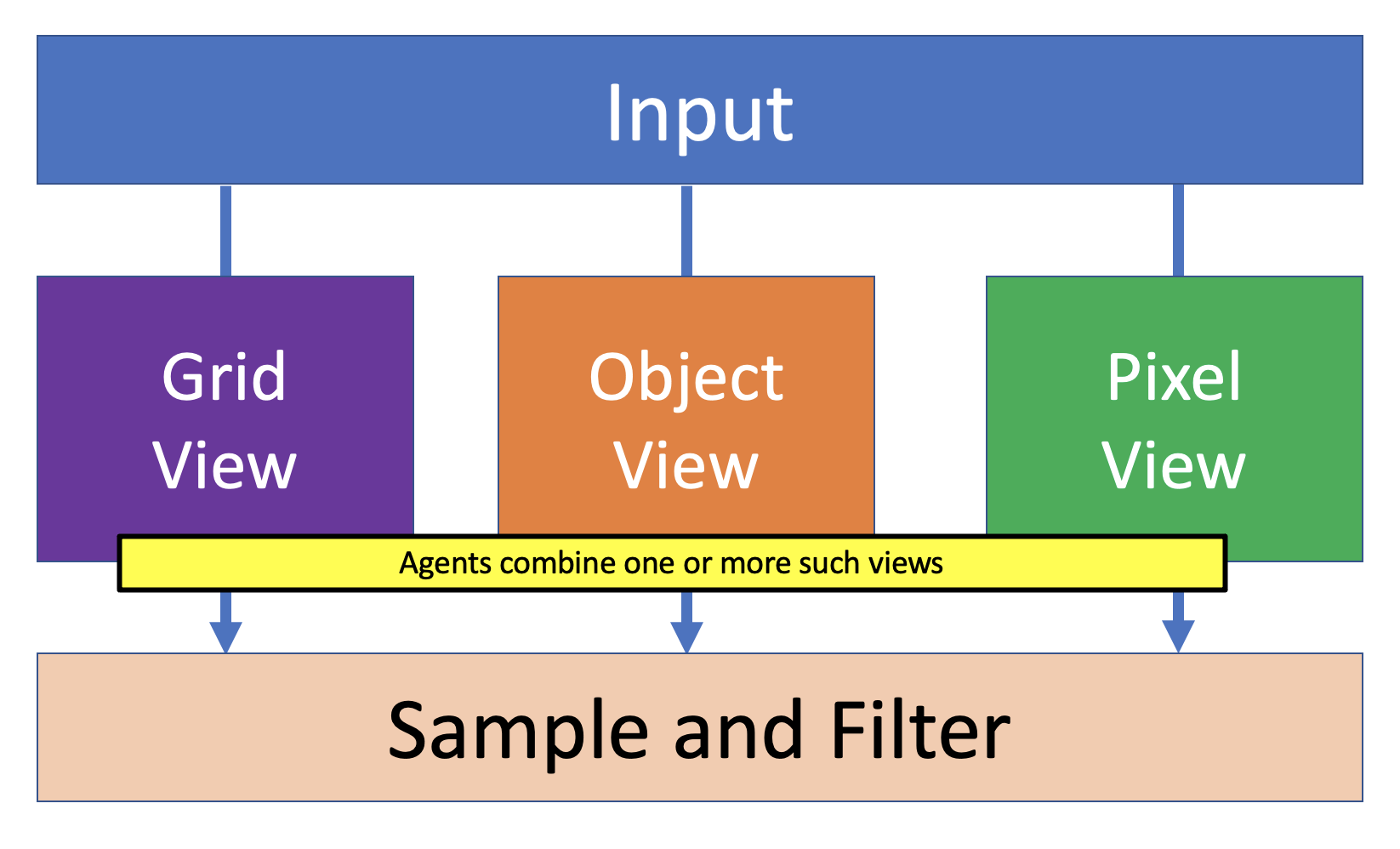}
\caption{The overall Mass Sampling and Filtering process with various expert agents}
\label{fig: Agent Views}
\vspace{-5mm}
\end{figure}

\noindent \textbf{Mass Sampling and Filtering.} Next, we use the OpenAI API for GPT-4 May 24 2023 version with a temperature of 0.7 to ensure a diverse range of outputs. We use the OpenAI API and the web browser interface for GPT-4 interchangeably. We employ a mass sampling and filtering process to generate code, much like in AlphaCode \cite{li2022competition} (see Fig. \ref{fig: Agent Views}). \textit{Grid View} is always there unless there is context length limitation. We can choose between toggling \textit{Object View} (10 types) and \textit{Pixel View} for the agents (at least one must be active), which leads to a total of $10\times2 = 20$ agents (See Appendix \ref{appendix: abstraction} for details). We utilise each expert agent three times each, with at most three feedback loop iterations, and filter the output codes which can solve the Task Demonstration to try it out on the Task Input. If there are multiple such codes, we randomly pick three to test it out. Any of these three solutions passing the Test Input will be counted as a solve, which is in line with the Kaggle competition and Lab 42's \href{https://lab42.global/arcathon/}{ARCathon}.

\begin{table*}[t]
\caption{Number of tasks solved, not solved and partially solved (Program works for Task Demonstration but not for Test Input/Output out of 111 Training Set tasks). See Appendix \ref{appendix: task details} for breakdown of tasks solved by each view type.}
\label{table: number of tasks solved}
\begin{center}
\begin{tabular}{|c|c|c|c|} 
\hline
\textbf{Total Tasks} & \textbf{Tasks Solved} & \textbf{Tasks Not Solved} & \textbf{Tasks Partially Solved}\\
 \hline
111 & 50 & 58 & 3\\
\hline
\end{tabular}
\end{center}
\end{table*}

\begin{table}[t]
\caption{Tasks not solved but with correct description}
\label{table: number of tasks correct description}
\begin{center}
\begin{tabular}{|c|c|} 
\hline
\textbf{Total Tasks Not Solved} & \textbf{Correct description}\\
 \hline
61 & 8 \\
\hline
\end{tabular}
\end{center}
\end{table}

\begin{table}[t]
\caption{Tasks solved with iterative feedback loop after either incorrect output or compile error}
\label{table: number of tasks feedback loop}
\begin{center}
\begin{tabular}{|c|c|c|} 
\hline
\textbf{Total} & \textbf{Incorrect Output} & \textbf{Compile Error}\\
 \hline
50 & 6 & 1\\
\hline
\end{tabular}
\end{center}
\end{table}

\section{Results}
\textbf{Overall.} Overall, as shown in Table \ref{table: number of tasks solved}, our method solves 50 out of 111 Training Set ARC tasks which could fit within the context length. This is about a 45\% solve rate, which is quite remarkable as the current ARC world record solve rate is 30.5\% (though this is on the hidden test set), according to
\url{https://lab42.global/arcathon/updates/}.

\noindent \textbf{Coding Issues.} To see how many of the unsolved problems are due to coding issues, we check how many of them have the correct description as evaluated by a human, but not have the correct code. This turns out to be 8 out of 61, as shown in Table \ref{table: number of tasks correct description} (See Appendix \ref{appendix: task details} for details). This means that if we could learn the primitive/helper functions better and have a wider range to choose from, we can improve solve rate. To solve the rest of the problems, we will have to incorporate better views - it is observed that GPT-4 cannot solve line continuation tasks, especially for diagonal lines, grid manipulation tasks, and symmetry tasks easily, and these could easily be incorporated as additional views.

\noindent \textbf{Iterative Feedback.} To see how much iterative environmental feedback helps, we look at number of tasks solved with the iterative environment feedback loop. This turns out to be 7 tasks out of 50, as shown in Table \ref{table: number of tasks feedback loop} (See Appendix \ref{appendix: task details} for details). This is quite significant, and highlights the importance of environmental feedback. 

\section{Discussion}

The results are promising, and GPT-4 agents with various combination of views can solve different types of problems well, as compared to just using the original \textit{Grid View}. It was also sometimes observed that \textit{Object View} had to go with \textit{Pixel View} for a consolidation of information across both views in order to solve the task. This reinforces the view that there should not be just one abstraction space, but multiple abstraction spaces which could be used in combination with each other.

Empirical observation has shown that GPT-4 with primitive function grounding can solve more tasks than without. It is a better way at encoding priors than with just text alone. Overall, GPT-4 is great at solving tasks which are made up of a combination of primitive functions.

It was observed that function names and descriptions are very important - GPT-4 tends to choose functions semantically similar to what it intends to do, and the changing of a function name to something irrelevant may cause it not to be used.

\section{Improvements}
GPT-4 agents cannot do tasks that have no relevant priors encoded in the primitive functions well, such as scaling of objects, symmetry, continuation of lines, overlay of grids with logical rules, grid manipulation like cropping, translating, changing of shape. Furthermore, it is weak when there is more than one relation, and this type of problems benefit from the iterative environment feedback loop. By setting the new input as the output that GPT-4's program outputs, it is in effect taking a step towards the solution and helps GPT-4 better associate the simpler input-output relationship. 

GPT-4 has been observed to use primitive functions not meant for the view, for example, \textit{Pixel View} Agent using the get\_objects function. Hence, giving too much context might affect performance. This is similar to \citet{xu2023llms} when the performance declined after adding in relations between objects. This reinforces our idea that it is best to split up into multiple expert agents with separate views and only relevant primitive functions.

Based on our experimental results, we propose new views/agents in Appendix \ref{appendix:insights}.

% \section{Ablation Studies}
% - Remove functional grounding
% - Remove chain of thought prompting altogether and ask to generate the program

\section{Future Work}
Currently, we use all agents in a brute-force manner for a task. In order to reduce computation (and cost), we could perhaps have a classifier which takes in previous examples as input to learn how to classify a new problem into a category, so that the right agents can be used to solve it.

Currently, the primitive functions are hand-engineered based on observation of the first 50 tasks in the training set, and are also not a complete set. We will try to incorporate a way for GPT-4 to be prompted to create new primitive functions, and add those successful functions which could solve a new task to the list of primitive functions, much like Voyager \cite{wang2023voyager}. One way is to add any transform\_grid function that is successful as a new primitive function, as long as the description of the function is different from existing ones.

\section{Conclusion}
Overall, LLMs as a system of multiple expert agents with environmental feedback is a promising approach towards solving the ARC Challenge. To facilitate further research using this approach, our code can be found at \url{https://github.com/tanchongmin/ARC-Challenge/}.

\section*{Acknowledgements}
This research is supported by the National Research Foundation, Singapore under its AI Singapore Programme (AISG Award No: AISG-GC-2019-002). Any opinions, findings and conclusions or recommendations expressed in this material are those of the author(s) and do not reflect the views of National Research Foundation, Singapore.

Many thanks for the various intelligent people who have encouraged me to pursue the GPT4 route to solve ARC or have provided valuable insights - Pascal Kaufmann, Rolf Pfister, Michael Hodel, Simon Strandgaard, Douglas Miles, Richard Cottrill, Leonard Tan and many others. If your name is not here, do not worry, you can be in the future paper improving on this work:)

\newpage
\bibliography{main.bib}
\bibliographystyle{icml2021}

\newpage
\appendix
\onecolumn

\newpage
\noindent {\bf APPENDIX}
~\newline

The appendix contains the following sections:
\begin{enumerate}[label=\Alph*]
    \item Full Prompt Details for GPT-4 - This details the entire prompt used for GPT-4
    \item Primitive Functions and Conditional Functions - This details all the primitive functions and conditional functions used for the grounding of GPT-4
    \item Abstraction Views - This details the various abstraction views of Grid, Object and Pixel
    \item GPT-4 Output Examples - This showcases GPT-4's output to the prompt
    \item Task Solved Details - This details the breakdown of the tasks solved by view type
    \item Proposed Agent Types - This details the proposed agent types which may help increase solve rate of GPT-4 for the ARC Challenge.
\end{enumerate}

~\newline

\newpage
\section{Full Prompt Details for GPT-4}
\label{appendix: prompt}

This section details the prompts used for GPT-4. The prompts are split up into a user prompt and a system prompt, as required by the GPT-4 API. If we do not use the API and use the web browser interface instead, we put the user prompt at the start of the prompt and the system prompt at the end of the prompt with the appropriate headings.

\subsection{User Prompt}

\begin{framed}
\noindent All coordinates are given as (row,col). Use get\_size(grid) to return (len(grid),len(grid[0])).\\
To get objects, use get\_objects(diag=False, by\_row=False, by\_col=False, by\_color=False, multicolor=False, more\_info=True) \# replace this with whatever object view was used

\noindent \textit{[JSON with various abstraction views of input/output, and input and output grid size]} \\
\textit{[Environmental Feedback - Either empty if first iteration, otherwise either Compile Error, or Output Error Message]}
\end{framed}

\subsection{Environmental Feedback (Compile Error Message)}
\begin{framed}
\noindent Previous Code: \textit{[Code]}\\
Error Message: \textit{[Error Message]}\\
Previous Overall Pattern: \textit{[Overall Pattern]}\\
Your code had compilation errors. Correct it.
\end{framed}

\subsection{Environmental Feedback (Output Error Message)}
If there is an output error (code generated output does not match Task Demonstration output), we treat this output as the new input and ask GPT-4 to get the new input-output relation.
\begin{framed}
\noindent Use the transform\_grid function to get the right relation from 'input' to 'output'
\end{framed}

\newpage
\subsection{System Prompt}
Refer to Appendix \ref{appendix: helper functions} for details for the helper/primitive and conditional functions.

\begin{framed}
\noindent You are given a series of inputs and output pairs. \\
The values from 'a' to 'j' represent different colors. '.' is a blank cell.\\
For example, [['.','a','.'],['.','.','b']] represents a 2 row x 3 col grid with color a at position (1,0) and color b at position (2,1).\\
Coordinates are 2D positions (row, col), row representing row number, col representing col number, with zero-indexing.\\
Input/output pairs may not reflect all possibilities, you are to infer the simplest possible relation.\\

\noindent 
\textit{[Helper/Primitive Functions Description + Example]}\\
\textit{[Conditional Functions Description + Example]} \\

\noindent You are to output the following in json format: \\
\{'reflection': 'reflect on the answer', \\
'pixel\_changes': 'describe the changes between the input and output pixels, focusing on movement or pattern changes', \\
'object\_changes': 'describe the changes between the input and output objects, focusing on movement, object number, size, shape, position, value, cell count', \\
'helper\_functions': 'list any relevant helper\_functions for this task', \\
'overall\_pattern': 'describe the simplest input-output relationship for all input-output pairs', \\
'program\_instructions': 'Plan how to write the python function and what helper functions and conditions to use', \\
'python\_program': "Python function named 'transform\_grid' that takes in a 2D grid and generates a 2D grid. Output as a string in a single line with $\backslash$n and $\backslash$t."\}.\\
Do not use quotation marks ' or " within the fields unless it is required for the python code
\end{framed}

\newpage
\section{Primitive Functions and Conditional Functions}
\label{appendix: helper functions}

This section details all the primitive functions and conditional functions used for GPT-4. Currently, these functions are not learnable, and are defined via tuning by a human over the first 50 ARC training tasks. In the future, we intend for these functions to be learned and expanded from a starting set independent of human interaction.

For more information of how these functions were implemented, refer to the Jupyter Notebook provided in \url{https://github.com/tanchongmin/ARC-Challenge}.

\newpage
\subsection{Primitive Functions}
This is the way we prompted GPT-4 to understand the format for the primitive (helper) functions. This is essentially the 
\begin{framed}
\noindent Each of the input-output relation can be done with one or more helper functions chained together. \\
Some relations require other functions, which you will need to come up with yourself.\\
Objects are tight-fitted grids (no empty row or column) with a top left coordinate, which can be used for easy manipulation of multiple coordinates.\\
You can create your own objects by just creating a dictionary with 'tl' and 'grid'\\
You can change an object's position by using 'tl' and its composition using 'grid'.\\
You should start each program by copying input grid or empty\_grid or crop\_grid of desired output size.\\
Then, fill the grid by using the fill helper functions.\\
If you use the fill functions with a '.' value, it is equivalent to removing parts of the grid.\\

Helper functions:\\
- get\_objects(grid,diag=False,by\_row=False,by\_col=False,by\_color=False,multicolor=False,more\_info = True): 
Takes in grid, returns list of object dictionary: top-left coordinate of object ('tl'), 2D grid ('grid')
by\_row views splits objects by grid rows, by\_col splits objects by grid columns, by\_color groups each color as one object, multicolor means object can be more than one color.
Empty cells within objects are represented as '\$'.
If more\_info is True, also returns size of grid ('size'), cells in object ('cell\_count'), shape of object ('shape')\\
- get\_pixel\_coords(grid): Returns a dictionary, with the keys the pixel values, values the list of coords, in sorted order from most number of pixels to least\\
- empty\_grid(row, col): returns an empty grid of height row and width col\\
- crop\_grid(grid, tl, br): returns cropped section from top left to bottom right of the grid\\
- tight\_fit(grid): returns grid with all blank rows and columns removed\\
- combine\_object(obj\_1, obj\_2): returns combined object from obj\_1 and obj\_2. if overlap, obj\_2 overwrites obj\_1\\
- rotate\_clockwise(grid, degree=90): returns rotated grid clockwise by a degree of 90, 180, 270 degrees\\
- horizontal\_flip(grid): returns a horizontal flip of the grid\\
- vertical\_flip(grid): returns a vertical flip of the grid\\
- replace(grid, grid\_1, grid\_2): replaces all occurences of grid\_1 with grid\_2 in grid\\
- get\_object\_color(obj): returns color of object. if multicolor, returns first color only\\
- change\_object\_color(obj, value): changes the object color to value\\
- fill\_object(grid, obj, align=False): fills grid with object. If align is True, makes grid same size as object\\
- fill\_row(grid, row\_num, value, start\_col=0, end\_col=30): fills output grid with a row of value at row\_num from start\_col to end\_col (inclusive)\\
- fill\_col(grid, col\_num, value, start\_row=0, end\_row=30): fills output grid with a column of value at col\_num from start\_row to end\_row (inclusive)\\
- fill\_between\_coords(grid, coord\_1, coord\_2, value): fills line between coord\_1 and coord\_2 with value\\
- fill\_rect(grid,tl,br,value): fills grid from tl to br with value. useful to create rows, columns, rectangles\\
- fill\_value(grid, pos, value): fills grid at position with value
\end{framed}

\newpage
This is the way we one-shot prompted GPT-4 for the primitive functions.
\begin{framed}
\noindent assert get\_objects([['a','a','a'],['a','.','a'],['a','a','a']],more\_info=False)==[\{'tl':(0,0),grid':[['a','a', 'a'],['a','.','a'],['a','a','a']]\} ,\{'tl':(1,1),'grid':[['\$']]\}]\\
assert get\_pixel\_coords([['a','a'],['d','f']])==\{'a':[(0, 0),(0, 1)],'d':[(1, 0)],'f':[(1, 1)]\}\\
assert empty\_grid(3, 2)==[['.','.'], ['.','.'], ['.','.']]\\
assert crop\_grid([['a','a','b'],['.','a','b']],(0, 0),(1, 1))==[['a','a'],['.','a']]\\
assert tight\_fit([['.','.','.'],['.','a','.'],['.','.','.']])==[['a']]\\
assert combine\_object(\{'tl':(0, 0),'grid':[['a','a'],['a','.']]\},\{'tl': (1, 1),'grid':[['f']]\})==\{'tl':(0, 0),'grid':[['a','a'],['a','f']]\}\\
assert rotate\_clockwise([['a','b'],['d','e']],90)==[['d','a'],['e','b']]\\
assert rotate\_clockwise([['a','b'],['d','e']],270)==[['b','e'],['a','d']]\\
assert horizontal\_flip([['a','b','c'],['d','e','f']])==[['c','b','a'], ['f','e','d']]\\
assert vertical\_flip([['a','b','c'],['d','e','f']])==[['d','e','f'],['a','b','c']]\\
assert replace([['a','.'],['a','a']],[['a','a']],[['c','c']])==[['a','.'],['c','c']]\\
assert change\_object\_color(\{'tl':(0,0),'grid':[['a','.']]\},'b')==\{'tl':(0,0),'grid':[['b','.']]\}\\
assert get\_object\_color(\{'tl':(0,0),'grid':[['a','.']]\})=='a'\\
assert fill\_object([['.','.'],['.','.']],\{'tl':(0, 1),'grid':[['c'],['c']]\})==[['.','c'],['.','c']]\\
assert fill\_value([['.','a'],['.','a']],(1,1),'b')==[['.','a'],['.','b']]\\
assert fill\_row([['a','a'],['c','a']],0,'b')==[['b','b'],['c','a']]\\
assert fill\_col([['a','a'],['c','a']],0,'b')==[['b','a'],['b','a']]\\
assert fill\_rect([['a','a'],['c','a']],(0,0),(1,1),'b')==[['b','b'],['b','b']]\\
assert fill\_between\_coords([['.','.']],(0,0),(0,1),'a')==[['a','a']]
\end{framed}

\newpage
\subsection{Conditional Functions}
This is the way we prompted GPT-4 to understand the format for the conditional functions. 
\begin{framed}
\noindent Each helper function can be conditional. \\
The conditions can be:\\
- by attribute, such as shape, color, position, size, cell number of object\\
- the condition can be an attribute on all objects, for instance, objects with the most common or least common value, or objects with the most or least common shape\\
- by position of pixels, such as row or column\\
- by neighbouring cell types or values\\

\noindent There are some conditional functions to help you.\\
- object\_contains\_color(obj, value): returns True/False if object contains a certain value\\
- on\_same\_line(coord\_1, coord\_2): Returns True/False if coord\_1 is on the same line as coord\_2. line\_type can be one of ['row', 'col', 'diag']
\end{framed}

This is the way we one-shot prompted GPT-4 for the conditional functions.
\begin{framed}
\noindent assert object\_contains\_color(\{'tl':(0,0),'grid':[['a']]\},'a')==True\\
assert on\_same\_line((1,1),(1,2),'row')==True\\
assert on\_same\_line((1,1),(2,1),'col')==True\\
assert on\_same\_line((1,1),(2,2),'diag')==True
\end{framed}

\newpage
\section{Abstraction Views}
\label{appendix: abstraction}

This section details how each abstraction view is represented.

\noindent \textbf{Grid View} - This provides the entire grid in ASCII format, in a 2D array. This is used when there are arbitrary patterns to find out. Note that we replace the initial JSON task input of grids with 0-9, with grids of '.','a','b','c','d','e','f','g','h','i'. This not only serves to prevent GPT-4 from trying to perform arithmetic on pixel values, but also eradicates the possibility that GPT-4 has seen the public training or evaluation set online, because it is of a different form.

\noindent \textbf{Object View} - This provides GPT-4 with an input view that groups contiguous cells of the same or different colour (non-zero cells) together and gives their attributes - top left coordinate, tight-fitted grid layout, shape, size, cell count. This is used when we would like to manipulate groups of related cells as one object.

Within object view, there are a few different types of categorising into objects. This is loosely inspired from Michael Hodel's DSL for ARC from \url{https://github.com/michaelhodel/arc-dsl}. We also have an option "more\_info", which helps with reducing context length when set to False. All in all, there are 10 different kinds of Object Views.

\begin{enumerate}[nosep,leftmargin=*]

    \item Attribute 1 - Colour (2 possibilities): Mono Colour vs Multi Colour - One way is to form objects is by grouping contiguous cells of the same connected either horizontally or vertically. The other way is to group contiguous cells of any colour except of colour '.' together. These two ways are termed "Mono Colour" and "Multi Colour" respectively.
    \item Attribute 2 - Constrains (5 possibilities): None/Row/Column/Colour/Diagonal - Row or column is used when the pattern is confined to just the row or column. Colour is used when we group objects by colour globally. It helps a lot to split a difficult problem into a smaller confined one where GPT-4 can perform association. Diagonal treats contiguous cells as diagonal connections as well.
    
\end{enumerate}

\noindent \textbf{Pixel View} - This provides GPT-4 with a view of the pixel value and all the corresponding coordinates, in a dictionary form with the pixel value as the key and the list of coordinates as the value. This is used when the input-output relation involves changing a group of pixels. This can also be used to supplement Object View. The benefit of using pixel view is that relations like shifting, adding or removing pixels at various positions are immediately obvious to GPT-4 as compared to Grid View. The downside though, is that pixel view is not robust to offsets of positions - if the starting grid is not tight fitted and does not start at (0,0), this can lead to problems with mapping.

\begin{figure}[h]
\centering
\includegraphics[width=0.2\textwidth]{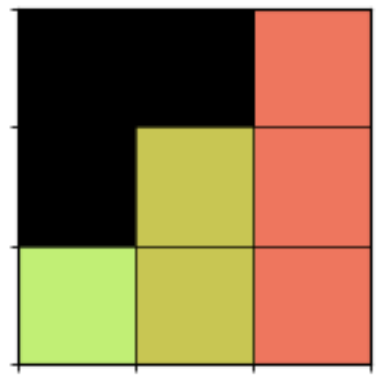}
\caption{A sample grid for an ARC Challenge Task - taken from Task Demonstration 1 of ARC Training Set d037b0a7}
\label{fig: Views for the ARC Puzzle}
\vspace{-5mm}
\end{figure}

\textbf{Example:} For Fig. \ref{fig: Views for the ARC Puzzle}, the abstraction spaces will be represented in text as:
\begin{enumerate}
    \item \textbf{Grid View:} [['.','.','f'],['.','d','f'],['c','d','f']]
    \item \textbf{Object View (Mono-Color):} [\{'tl':(0,2), 'grid':[['f'],['f'],['f']], 'size':(3,1), 'cell\_count':3, 'shape':[['x'],['x'],['x']]\}, 
    \{'tl':(1,1), 'grid':[['d'],['d']], 'size':(2,1), 'cell\_count':2, 'shape':[['x'],['x']]\},
    \{'tl':(2,0), 'grid':[['c']], 'size':(1,1), 'cell\_count':1, 'shape':[['x']]\}]
    \item \textbf{Pixel View:} \{'f':[(0,2),(1,2),(2,2)], 'd':[(1,1),(2,1)], 'c':[(2,0)]\}
\end{enumerate}

\newpage
\section{GPT-4 Output Examples}
\label{appendix: output examples}
In this section, we showcase GPT-4's output for a solved problem, and the output for an unsolved problem.

\subsection{GPT-4's Output for Solved Problems}

\begin{figure}[h]
\centering
	\begin{minipage}[t]{\textwidth}
		\includegraphics[width=\textwidth]{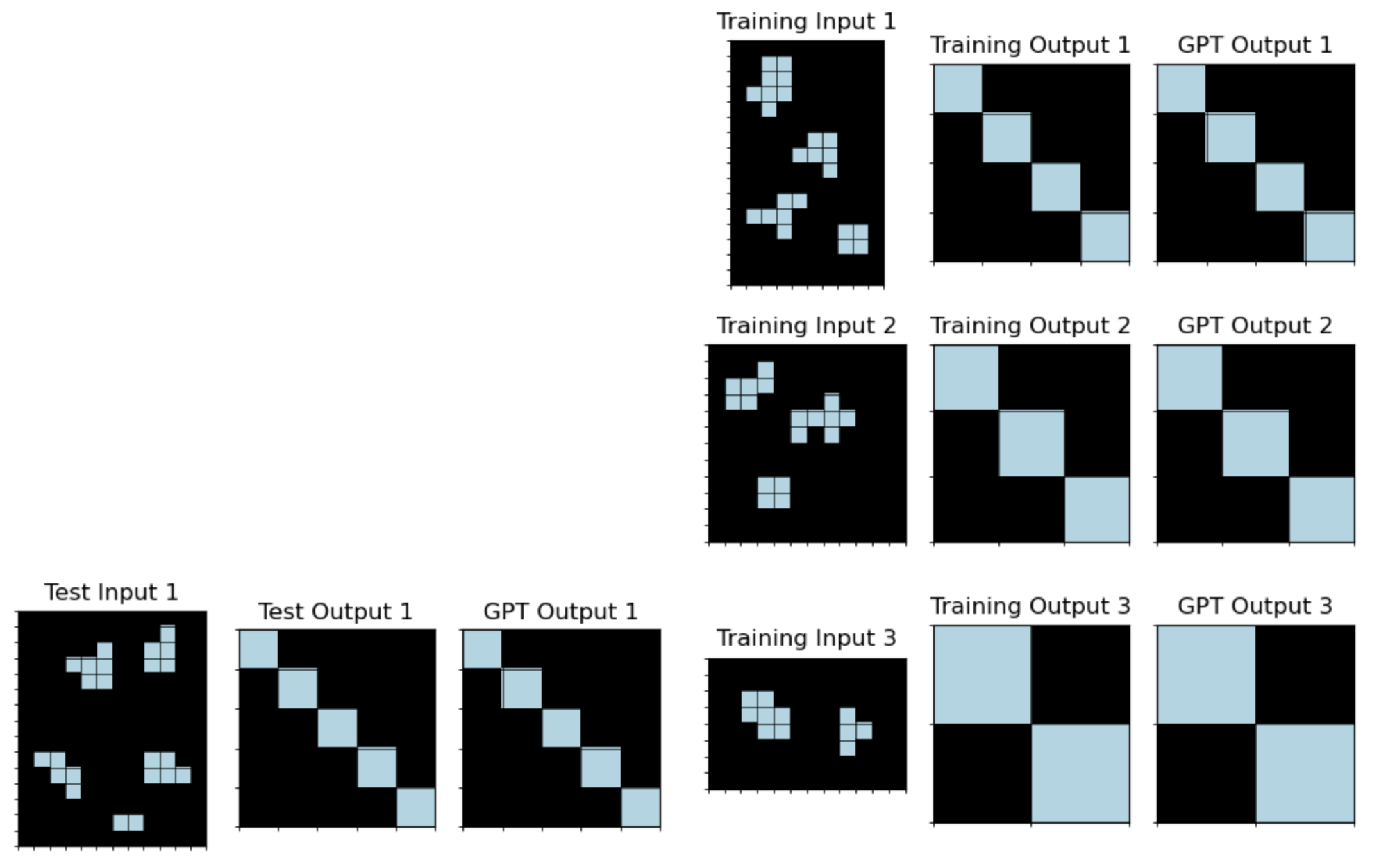}
	\end{minipage}%
	\caption{Training and Test Input/Output pairs for a Solved Problem, d0f5fe59, using Object View}
        \label{fig: solved workflow}
\end{figure}

\newpage
This is the JSON output of GPT-4 formatted into key-value pairs, and the "python\_program" provided correctly solves the Test Input.
\begin{framed}
\textbf{reflection:} This task involves objects found in the grid which are simplified and moved to a certain pattern in the output grid. The input grids contain a variety of object shapes, some of which are isolated pieces while others are complex shapes. These objects, regardless of their original form, are reduced to a single cell in the output grid and relocated to the diagonal line from the top-left to the bottom-right.

\textbf{pixel\_changes:} In the transformation from input to output, the number of pixels with color 'h' decreases. This decrease occurs because every 'h' color object is condensed to a single pixel in the output. Also, the pixels are relocated, following a diagonal line from top left to bottom right, where the first object is placed at (0,0), the second at (1,1), and so on, regardless of their initial positions.

\textbf{object\_changes:} The number of objects remains the same from input to output. However, their size, shape, and positions change significantly. Each object, regardless of its initial shape and size, is simplified to a single pixel in the output grid. The position of the objects follows a specific pattern in the output: they are relocated to the diagonal line from the top left to the bottom right. The sequence of placement on the diagonal line is according to the order of the objects in the input grid.

\textbf{helper\_functions:} The relevant helper functions for this task are get\_objects() for identifying objects in the input grid, fill\_value() for placing single pixels at specific locations in the output grid, and empty\_grid() for initializing the output grid.

\textbf{overall\_pattern:} The task takes an input grid containing a number of objects, simplifies each object into a single cell, and repositions these cells along a diagonal line from the top left to the bottom right. The order of placement on the diagonal is according to the order of the objects in the input grid. The size of the output grid is determined by the number of objects in the input grid.

\textbf{program\_instructions:} 1. Use get\_objects() function to retrieve all objects from the input grid. 2. Create an empty grid using empty\_grid(), where the number of rows and columns equals the number of objects retrieved. 3. Loop through each object obtained from get\_objects(). For each object, fill a cell in the new grid with the object color at a diagonal position corresponding to the index of the object in the objects list. This is done by calling fill\_value() with the index as both the row and column position.

\textbf{python\_program:} def transform\_grid(grid): objects = get\_objects(grid) new\_grid = empty\_grid(len(objects), len(objects)) for i, obj in enumerate(objects): new\_grid = fill\_value(new\_grid, (i, i), get\_object\_color(obj)) return new\_grid

\end{framed}

\newpage
\subsection{GPT-4's Output for Unsolved Problems}

\begin{figure}[h]
\centering
	\begin{minipage}[t]{\textwidth}
		\includegraphics[width=\textwidth]{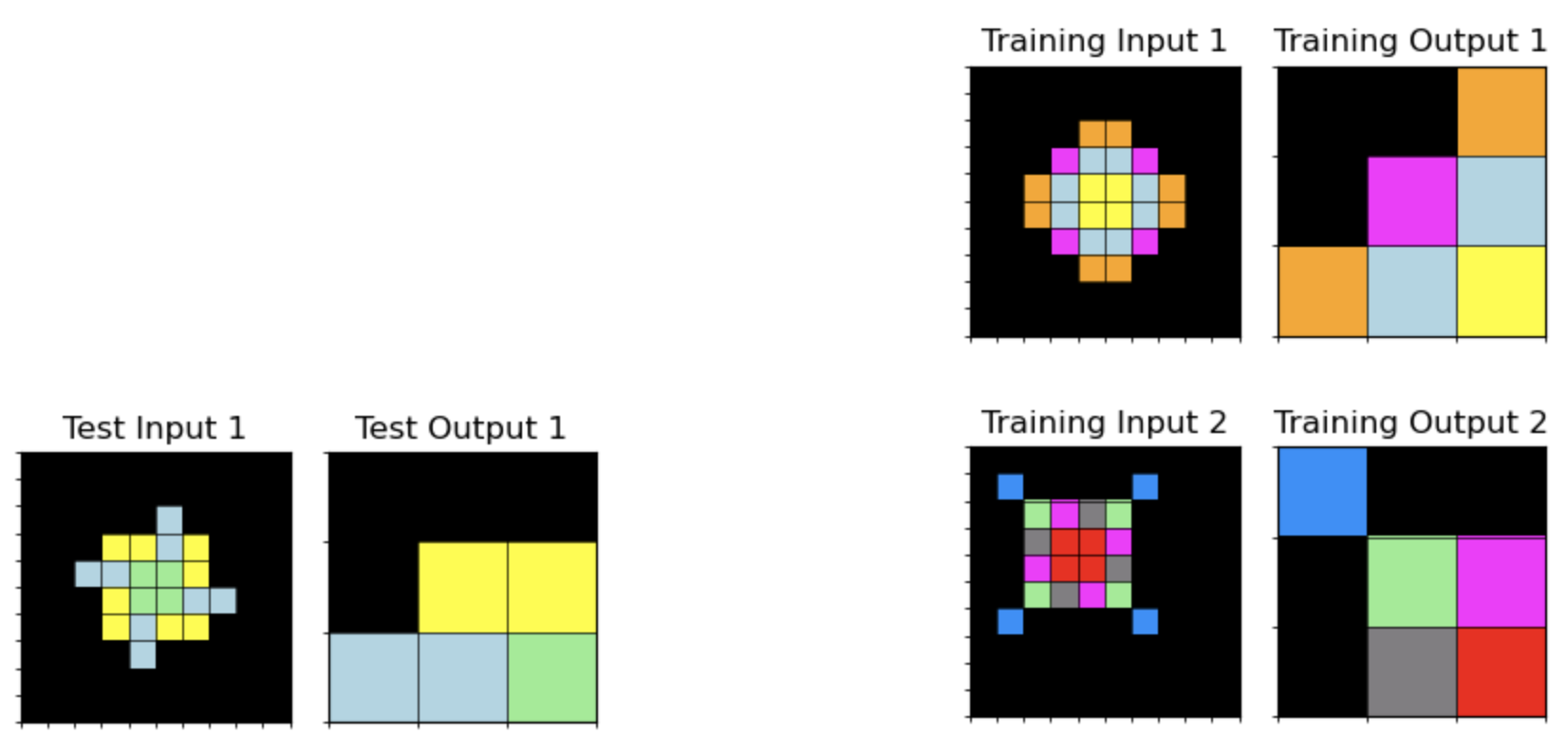}
	\end{minipage}%
	\caption{Training and Test Input/Output pairs for Iteration 1 for an unsolved Problem, 2013d3e2, using Object View. There was a compile error.}
        \label{fig: solved workflow}
\end{figure}

\begin{figure}[h]
\centering
	\begin{minipage}[t]{\textwidth}
		\includegraphics[width=\textwidth]{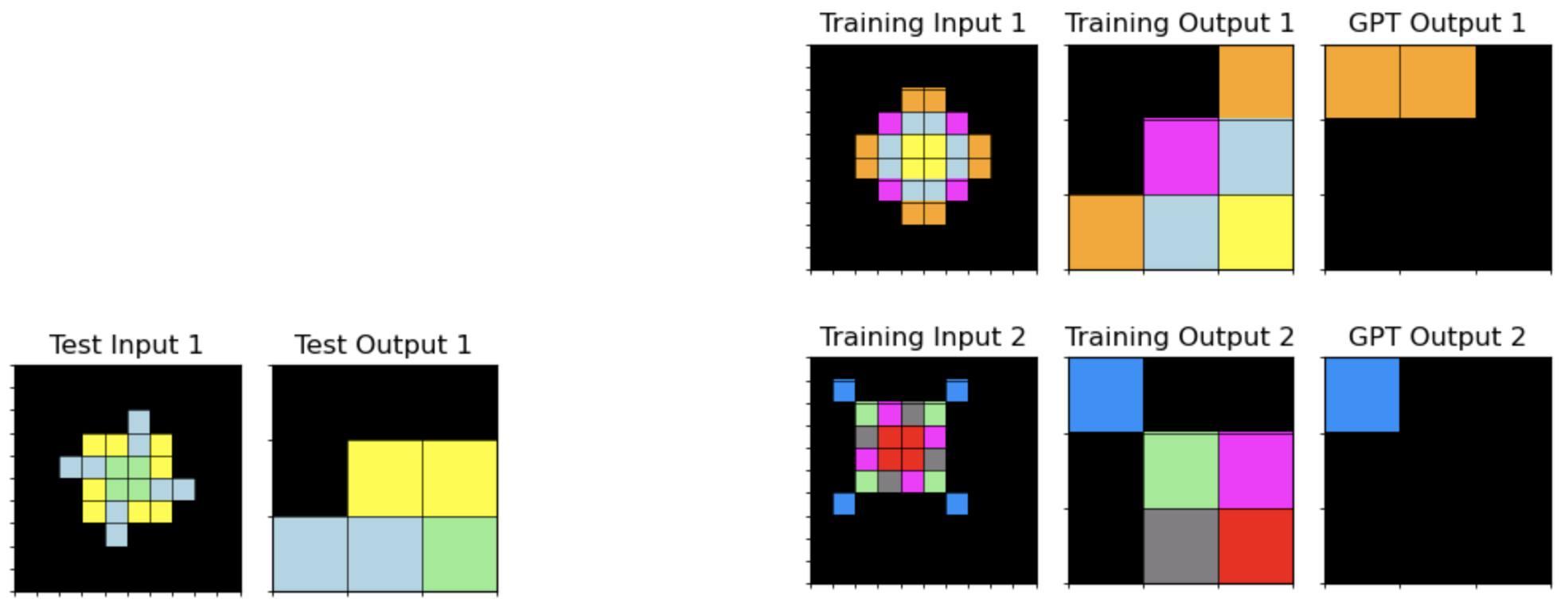}
	\end{minipage}%
	\caption{Training and Test Input/Output pairs for Iteration 2 for an unsolved Problem, 2013d3e2, using Object View. There was an incorrect program output. The run terminated at this iteration.}
        \label{fig: solved workflow}
\end{figure}

\newpage
This is the JSON output of GPT-4 formatted into key-value pairs, for Iteration 1 for 2013d3e2 using Object View, and the "python\_program" had a compile error. This is because it tried to crop from the object grid from the center rather than the top left of the object. It also failed to set diagonal = True and multicolor = True in the get\_object settings.
\begin{framed}
\textbf{reflection:} The task involves identifying a specific part of an object, cropping it, and outputting it. The specific part of the object is always three rows high and starts from the third column from the left. It's interesting to note that although the objects in the inputs are of different colors and shapes, the part that needs to be cropped is always at the same location.

\textbf{pixel\_changes:} The pixels in the output are a subset of the pixels from the input. They are always the pixels from the third, fourth, and fifth rows of the object, starting from the third column from the left. The colors of the pixels remain unchanged.

\textbf{object\_changes:} The size of the object is significantly reduced in the output. It goes from a 6x6 square in the input to a 3x3 square in the output. The shape of the object changes as a result of the cropping, but the relative positions of the pixels within the output object remain the same as in the input object.

\textbf{helper\_functions:} The following helper functions would be relevant for this task: get\_objects, crop\_grid, and fill\_object.

\textbf{overall\_pattern:} The task requires identifying the single object in the input, cropping a 3x3 square from it starting from the third column from the left, and outputting the cropped section. The part of the object that needs to be cropped is always the same and does not depend on the colors or the shape of the object.

\textbf{program\_instructions:} 1. Start with an empty grid of size 3x3. 2. Use the get\_objects function to identify the object in the input. 3. Crop a 3x3 square from the object starting from the third column from the left using the crop\_grid function. 4. Use the fill\_object function to fill the empty grid with the cropped section.

\textbf{python\_program:} 
def transform\_grid(grid): output = empty\_grid(3, 3) objects = get\_objects(grid) object = objects[0] cropped = crop\_grid(object['grid'], (2, 2), (4, 4)) fill\_object(output, \{'tl': (0, 0), 'grid': cropped\}) return output
\end{framed}

\newpage
This is the JSON output of GPT-4 formatted into key-value pairs, for Iteration 2 for 2013d3e2 using Object View, and the "python\_program" produced an incorrect output. It managed to solve the earlier compile error by using a try-catch block and just fill the first object's grid if there is any error. The program is still wrong because it decided to crop the object grid from the center rather than from the top left corner. It also failed to set diagonal=True and multicolor = True in the get\_object settings.

\begin{framed}
\textbf{reflection:} The task involves cropping a section from the input grid and outputting that section as the output grid. The error suggests that the cropping area defined in the function doesn't match the actual area in the object, which results in an out of range index error.

\textbf{pixel\_changes:} The pixels in the output are a subset of the pixels in the input. A 3x3 section from the center of the only object in the input grid is extracted and used to form the output grid.

\textbf{object\_changes:} The object in the output grid is a cropped section from the object in the input grid. There are no changes in the values of the pixels, the object simply reduces in size to a 3x3 grid. The top left cell of the output object corresponds to the cell in the third row and third column of the input object.

\textbf{helper\_functions:} The helper functions used for this task are get\_objects, crop\_grid, and fill\_object. get\_objects is used to get the object from the input grid. crop\_grid is used to crop a 3x3 section from the input object. fill\_object is used to fill the output grid with the cropped object.

\textbf{overall\_pattern:} The overall pattern is to identify the only object in the input grid, crop a 3x3 section from it starting from the third row and third column, and use this cropped section as the output grid.

\textbf{program\_instructions:} Start by copying the input grid. Use get\_objects to get the object from the input grid. Crop a 3x3 section from the input object starting from the third row and third column. Create an empty grid of size 3x3. Fill this grid with the cropped object and return it as the output. The python function will have to handle any exception to avoid out of range index error when the object is smaller than expected.

\textbf{python\_program:} def transform\_grid(grid): output = empty\_grid(3, 3) objects = get\_objects(grid) object = objects[0] try: cropped = crop\_grid(object['grid'], (2, 2), (4, 4)) except IndexError: cropped = object['grid'] fill\_object(output, {'tl': (0, 0), 'grid': cropped}) return output
\end{framed}

\newpage
\section{Task Solved Details}
\label{appendix: task details}

In this section, we provide a detailed breakdown of tasks solved in the main paper. For more details, check out the .html files provided in the \href{https://tinyurl.com/yet363w8}{Supplementary Materials}.\\

\begin{figure}[h]
\centering
	\begin{minipage}[t]{0.5\textwidth}
		\includegraphics[width=\textwidth]{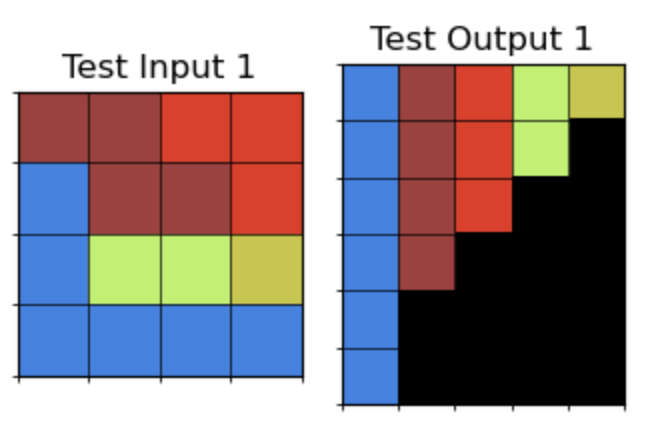}
	\end{minipage}%
	\hfill
	\begin{minipage}[t]{0.5\textwidth}
		\includegraphics[width=\textwidth]{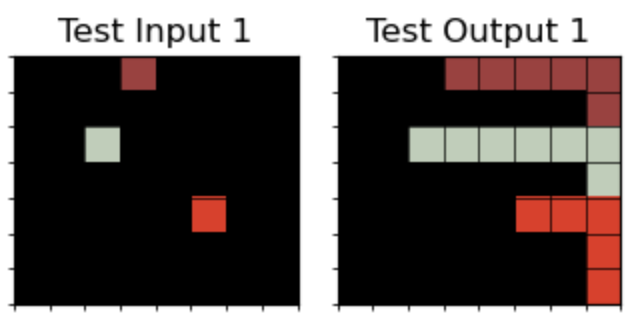}
	\end{minipage}
	\caption{Examples of tasks solved using only Pixel View. They can be used for a lot of situations and usually are the only performant view when the mapping is by colour, especially for irregular shape mappings.}
        \label{fig: pixel view}
\end{figure}

\begin{figure}[h]
\centering
	\begin{minipage}[t]{0.5\textwidth}
		\includegraphics[width=\textwidth]{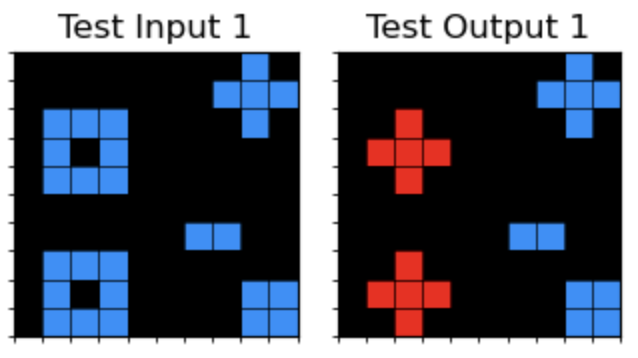}
	\end{minipage}%
	\hfill
	\begin{minipage}[t]{0.5\textwidth}
		\includegraphics[width=\textwidth]{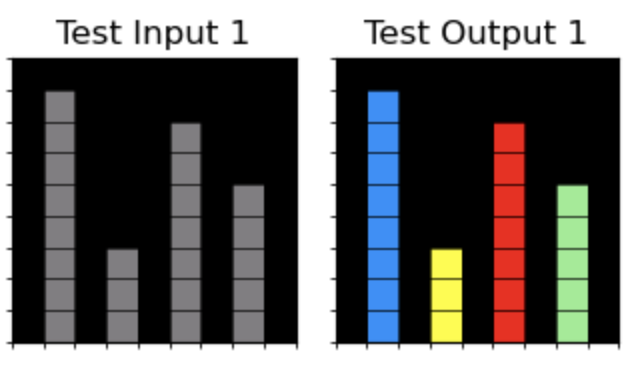}
	\end{minipage}
	\caption{Examples of tasks solved using only Object View. They are usually tasks which involve changing attribute of an object (e.g. shape, size, colour) or ranking object by attributes (e.g. size, cell count).}
        \label{fig: object view}
\end{figure}

\begin{figure}[h]
\centering
	\begin{minipage}[t]{0.5\textwidth}
		\includegraphics[width=\textwidth]{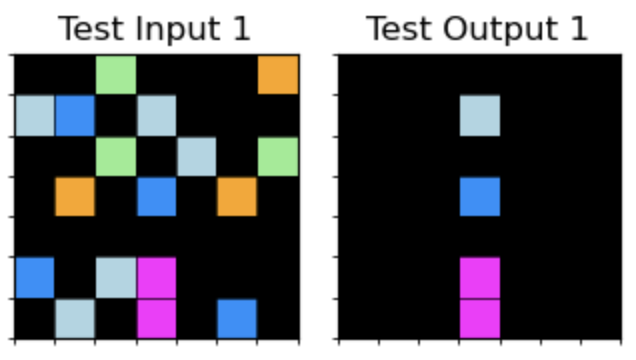}
	\end{minipage}%
	\caption{Examples of task solved using both Object View and Pixel View. They are usually tasks whereby we need to know the coordinates of the objects. Here, the objects are split according to columns. The task here is to just preserve the central column of pixels.}
        \label{fig: object and pixel view}
\end{figure}

\begin{figure}[h]
\centering
	\begin{minipage}[t]{0.5\textwidth}
		\includegraphics[width=\textwidth]{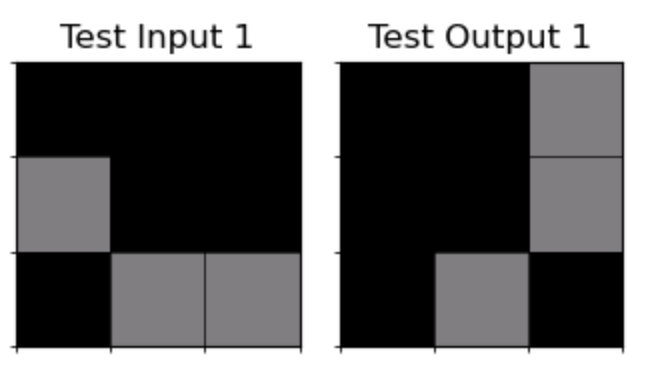}
	\end{minipage}%
	\hfill
	\begin{minipage}[t]{0.5\textwidth}
		\includegraphics[width=\textwidth]{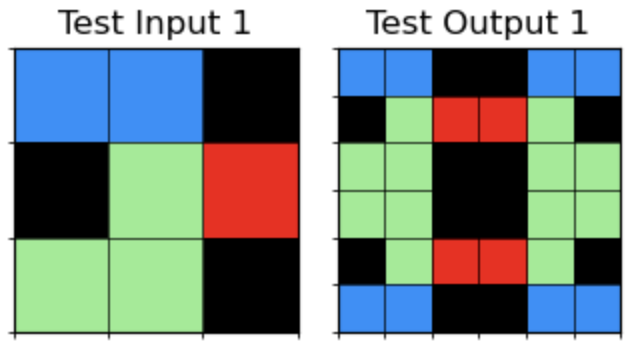}
	\end{minipage}
	\caption{Examples of tasks solved using only Grid View. They are mainly rotation and reflection tasks.}
        \label{fig: grid view}
\end{figure}

\noindent Table \ref{table: tasks solved agent types} shows the breakdown of the view types used to solve the 50 problems. As we can see, a variety of views are required, of which, the most frequently used one is Object View with 23 uses. Pixel View is the next most common with 19 uses. There is 1 task (d23f8c26) which used both Object View and Pixel View. There are some tasks for which Grid View alone is the best - 7 of them. Using only one single view is bound to get poorer results than using a combination of all these views. Some examples of tasks solved by the various views are given in Figs. \ref{fig: object view}, \ref{fig: pixel view}, \ref{fig: object and pixel view} and \ref{fig: grid view}.

\begin{table}[h]
\caption{Number of Tasks solved by each View Type. Grid View is used by default unless there is a limitation of the token length, in which case, it is toggled off}
\label{table: tasks solved agent types}
\begin{center}
\begin{tabular}{|c|c|} 
\hline
\textbf{View Type} & \textbf{Number of Tasks Solved}\\
 \hline
Total & 50\\
Object View & 23\\
Pixel View & 19\\
Object \& Pixel View & 1\\
No Object \& Pixel View (only Grid View) & 7\\
\hline
\end{tabular}
\end{center}
\end{table}

\newpage

\newpage
\noindent \textbf{Tasks Not Solved But with Overall Description Right.} List of tasks which are not solved but have overall description right:
\begin{enumerate}
    \item ea786f4a
    \item 88a62173
    \item 746b3537
    \item 321b1fc6
    \item 9565186b
    \item a3df8b1e
    \item f25ffba3
    \item 60b61512
\end{enumerate}

\begin{figure}[h]
\centering
	\begin{minipage}[t]{0.5\textwidth}
		\includegraphics[width=\textwidth]{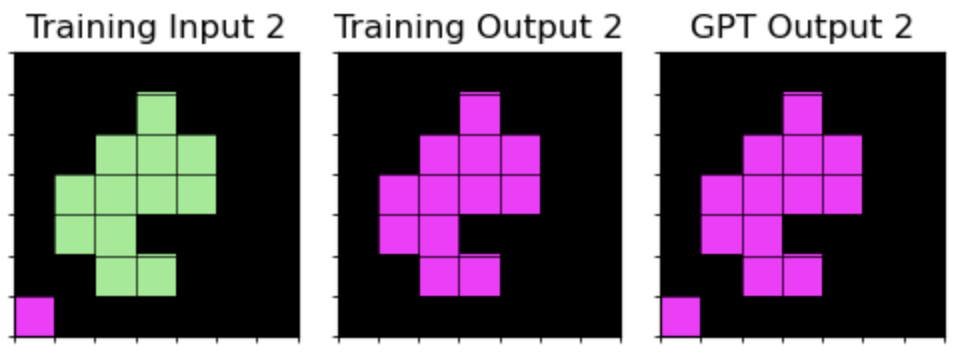}
	\end{minipage}%
	\hfill
	\begin{minipage}[t]{0.5\textwidth}
		\includegraphics[width=\textwidth]{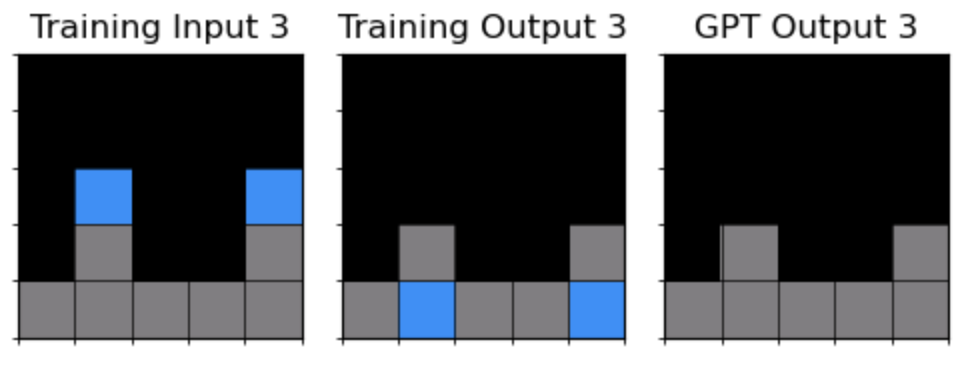}
	\end{minipage}
	\caption{Examples of tasks solved using Iterative Environment Feedback after incorrect output. They are usually tasks which involve multiple different steps, for example for the left task, changing colour of one object and removing another object, and for the right task, reducing height of the column and shifting the blue pixel to the bottom of the column.}
        \label{fig: iterative feedback}
\end{figure}

\newpage
\noindent \textbf{Tasks solved with Iterative Environment Feedback after Incorrect Output.} See Fig. \ref{fig: iterative feedback} for examples of such tasks. List of tasks which are solved over two iterated steps:
\begin{enumerate}
    \item aabf363d
    \item 496994bd
    \item 3618c87e
    \item 794b24be
    \item 3c9b0459
    \item ed36ccf7
\end{enumerate}

\begin{figure}[h]
\centering
	\begin{minipage}[t]{0.5\textwidth}
		\includegraphics[width=\textwidth]{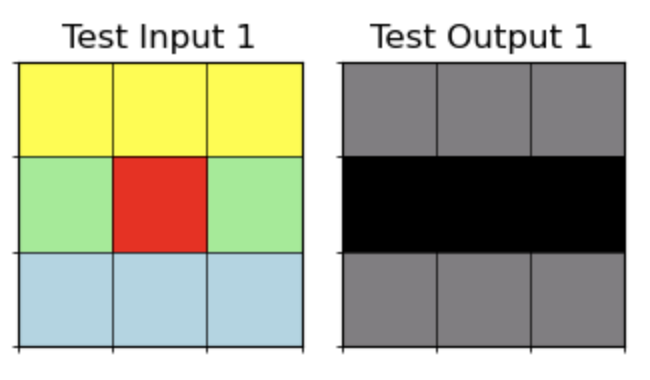}
	\end{minipage}%
	\caption{Examples of task solved using Iterative Environment Feedback after compile error. They are usually tasks which invoke a function that is not well explained in the prompt. Here, the program tried to do "output = copy(grid)" which is in response to the prompt "You should start each program by copying input grid or empty\_grid or crop\_grid of desired output
size". We could eliminate this error by simply elaborating the method to use in the prompt, for example, "You should start each program by copying input grid using copy.deepcopy(grid)".}
        \label{fig: compile error}
\end{figure}

\noindent \textbf{Tasks solved with Iterative Environment Feedback after Compilation Error.} See Fig. \ref{fig: compile error} for examples of such tasks. List of tasks which are solved after compilation error fixed:
\begin{enumerate}
    \item 25d8a9c8
\end{enumerate}

\newpage
\section{Proposed Agent Types}
\label{appendix:insights}
This section details more of the proposed new agent types which may help increase the solve rate of GPT-4 for the ARC Challenge.

\noindent \textbf{Better Object View.} We can perhaps do better by infusing more coordinates into the objects if we have context length. These coordinates would enable GPT-4 to know exactly which cell the object is positioned on. This was omitted from \textit{Object View} because for most object-related tasks, only the top left coordinate was necessary for translation, and so by omitting this we can reduce the context length. However, having the full coordinate view could be helpful, especially when comparing the change of one object to another. This can be seen when some tasks cannot be solved by \textit{Object View} or \textit{Pixel View} alone, but must be solved with a combination of both \textit{Object View} and \textit{Pixel View}.

\noindent \textbf{Object Relations View.} Relations between objects is something that could be improved. Right now, GPT-4 does not have the relative positions of all objects from each other, and from the edges and corners of the grid. This may be helpful in some tasks.

\noindent \textbf{Edge Detection View.} We note that GPT-4 does not process shapes naturally, and is lacking some idea of what a corner of the shape is, or what is the intersection of the shape. This may be possible if we give it a view of which are the corners of the object, which are the intersections, and this can help it to understand the 2D grid better. This is akin to giving it some form of processing through a Convolutional Neural Network (CNN) type network.

\noindent \textbf{Symmetry View.} We observe that GPT-4 does not have a good understanding of symmetry and fails to solve most tasks involving symmetry, especially rotational symmetry. Hence, imbuing this view as well as relevant helper/primitive functions could help with this kind of problems.

\noindent \textbf{Difference View.} We note that finding the differences between the input and output grid is key to solving ARC tasks. Perhaps one way of grounding GPT-4 to focus on the differences between input and output apart from prompting it with text, is to explicitly calculate the differences between the input and output over various aspects, and only show GPT-4 the ones that are different. This reduces context length and could potentially help GPT-4 to attend to these differences better. This is not strictly necessary as any good enough pattern associator can already deduce what the differences and similarities are with just the raw attributes without prompting - but this could perhaps nudge the answer towards the correct one more easily. The human brain also actively seeks out differences and that is how we learn, so having something to highlight differences could perhaps improve the system better.

\noindent \textbf{Diagonal View.} Being a token-based predictor, GPT-4 is actually quite weak at predicting diagonals since 2D representation is not naturally imbued in tokens. In fact, we observe that GPT-4 does row prediction better than both column and diagonal prediction. As such, it would be better if we could frame any problem as a row-based problem, including diagonals. One way to do this will be to rotate the grid 45 degrees so that diagonals become rows and columns, do manipulation on this grid, and then rotate it back -45 degrees so as to get back the original input grid.
%%%%%%%%%%%%%%%%%%%%%%%%%%%%%%%%%%%%%%%%%%%%%%%%%%%%%%%%%%%%%%%%%%%%%%%%%%%%%%%
%%%%%%%%%%%%%%%%%%%%%%%%%%%%%%%%%%%%%%%%%%%%%%%%%%%%%%%%%%%%%%%%%%%%%%%%%%%%%%%

\end{document}